\documentclass[conference]{IEEEtran}
\IEEEoverridecommandlockouts
\makeatletter
\def\ps@IEEEtitlepagestyle{%
  \def\@oddfoot{\mycopyrightnotice}%
  \def\@oddhead{\hbox{}\@IEEEheaderstyle\leftmark\hfil\thepage}\relax
  \def\@evenhead{\@IEEEheaderstyle\thepage\hfil\leftmark\hbox{}}\relax
  \def\@evenfoot{}%
}
\def\mycopyrightnotice{%
  \begin{minipage}{\textwidth}
  \centering \scriptsize
  Copyright~\copyright~2024 IEEE. Personal use of this material is permitted. Permission from IEEE must be obtained for all other uses, in any current or future media, including\\reprinting/republishing this material for advertising or promotional purposes, creating new collective works, for resale or redistribution to servers or lists, or reuse of any copyrighted component of this work in other works by sending a request to pubs-permissions@ieee.org.
  \end{minipage}
}
\makeatother

\usepackage{cite}
\usepackage{amsmath,amssymb,amsfonts}
\usepackage{algorithmic}
\usepackage{graphicx}
\usepackage{textcomp}
\usepackage[]{algorithm2e}

\makeatletter
\newcommand{\linebreakand}{%
  \end{@IEEEauthorhalign}
  \hfill\mbox{}\par
  \mbox{}\hfill\begin{@IEEEauthorhalign}
}
\makeatother

\usepackage{xcolor}
\def\BibTeX{{\rm B\kern-.05em{\sc i\kern-.025em b}\kern-.08em
    T\kern-.1667em\lower.7ex\hbox{E}\kern-.125emX}}
\begin{document}

\title{EndToEndML: An Open-Source End-to-End Pipeline for Machine Learning Applications
}

\author{\IEEEauthorblockN{1\textsuperscript{st} Nisha Pillai}
\IEEEauthorblockA{\textit{Mississippi State University} \\
\textit{Mississippi State, USA}\\
pillai@cse.msstate.edu}
*Corresponding author
~\\
\and
\IEEEauthorblockN{2\textsuperscript{nd} Athish Ram Das}
\IEEEauthorblockA{\textit{Mississippi State University} \\
\textit{Mississippi State, USA}\\
ar2903@msstate.edu}

~\\
\and
\IEEEauthorblockN{3\textsuperscript{rd} Moses Ayoola}
\IEEEauthorblockA{\textit{Mississippi State University} \\
\textit{Mississippi State, USA}\\
mba185@msstate.edu}
~\\ 
\linebreakand
\IEEEauthorblockN{4\textsuperscript{th} Ganga Gireesan}
\IEEEauthorblockA{\textit{Mississippi State University} \\
\textit{Mississippi State, USA}\\
gg733@msstate.edu}
~\\
\and
\IEEEauthorblockN{5\textsuperscript{th} Bindu Nanduri}
\IEEEauthorblockA{\textit{Mississippi State University} \\
\textit{Mississippi State, USA}\\
bnanduri@cvm.msstate.edu}
~\\
\and
\IEEEauthorblockN{6\textsuperscript{th} Mahalingam Ramkumar}
\IEEEauthorblockA{\textit{Mississippi State University} \\
\textit{Mississippi State, USA}\\
ramkumar@cse.msstate.edu}
}
\maketitle

\begin{abstract}
Artificial intelligence (AI) techniques are widely applied in the life sciences. However, applying innovative AI techniques to understand and deconvolute biological complexity is hindered by the learning curve for life science scientists to understand and use computing languages. An open-source, user-friendly interface for AI models, that does not require programming skills to analyze complex biological data will be extremely valuable to the bioinformatics community. With easy access to different sequencing technologies and increased interest in different ‘omics’ studies, the number of biological datasets being generated has increased and analyzing these high-throughput datasets is computationally demanding. The majority of AI libraries today require advanced programming skills as well as machine learning, data preprocessing, and visualization skills. In this research, we propose a web-based end-to-end pipeline that is capable of preprocessing, training, evaluating, and visualizing machine learning (ML) models without manual intervention or coding expertise. By integrating traditional machine learning and deep neural network models with visualizations, our library assists in recognizing, classifying, clustering, and predicting a wide range of multi-modal, multi-sensor datasets, including images, languages, and one-dimensional numerical data, for drug discovery, pathogen classification, and medical diagnostics.
\end{abstract}


\section{Introduction}
Biomedical research communities are increasingly in need of more accessible and interpretable artificial intelligence (AI) tools to extract insights from increasingly large and complex datasets. Performing machine learning on multifaceted biological data currently requires specialized programming expertise, limiting its adoption to bioinformaticians and computational biologists alone. The development of an open-source, user-friendly interface that enables researchers without coding skills to apply AI algorithms to data such as the microbiome related to soil, livestock, plant health, and other domains could dramatically expand the user base. By abstracting away the coding complexity through an intuitive graphical interface and pre-built analysis workflows, more biologists could utilize the predictive and pattern recognition capabilities of AI. This could accelerate discovery and the generation of hypotheses from high-dimensional biological data. Additional capabilities like interactive visualizations could also improve user understanding of the patterns and relationships detected by the models. An open-source platform would facilitate collaboration and allow customization of the built-in AI tools as methodologies continue to evolve. Overall, democratizing AI would equip a wider range of biologists to derive clinically and biologically relevant insights from increasingly complex and multi-modal data. This can profoundly enhance and accelerate biomedical research.

Nowadays, there exists a wide array of machine learning libraries and frameworks, such that even seasoned practitioners have difficulty discerning the appropriate tools given each solution's narrow capability. Furthermore, integration of these disjointed components is cumbersome when attempting to bring end-to-end functionality together. The greatest potential for impactful applications of machine learning lies in interdisciplinary partnerships. These synergies could be enhanced through a universal framework accessible to non-specialists. Placing straightforward yet potent machine learning capabilities into more hands accelerates innovation by allowing users of any skill level to move ML from promise to practical solutions.

\begin{figure}[htbp]
\centering
\includegraphics[width=8cm]{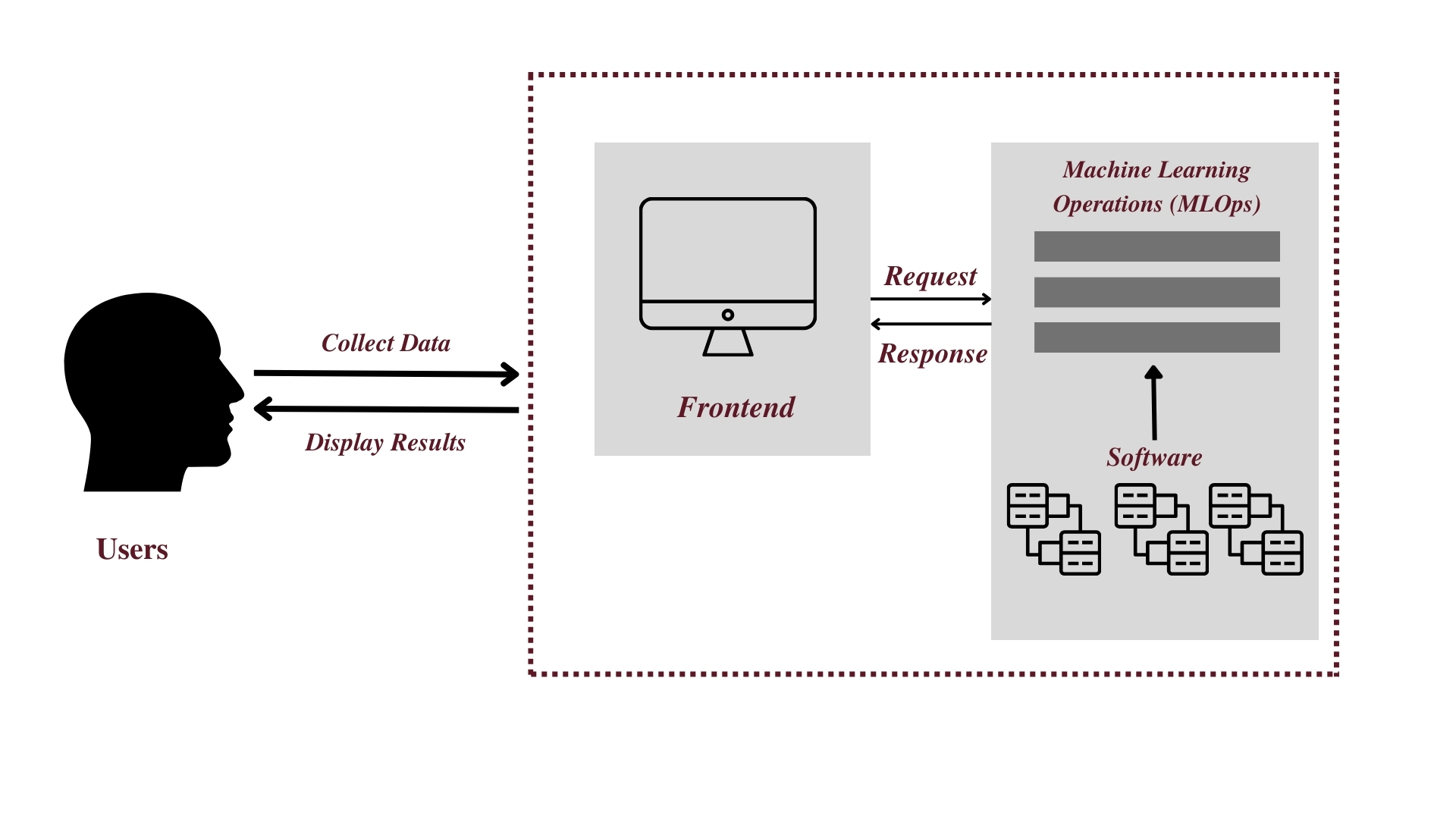}% This is a *.eps file
\caption{\centering Overall architecture of EndToEndML}
\label{fig:1}
\end{figure}

As an integrated solution to bridge expertise gaps hampering ML adoption, we have developed EndToEndML, an easy-to-use web application suite spanning the machine learning pipeline from data preprocessing to model deployment. Using an intuitive graphical interface that requires no programming knowledge, users can upload datasets and produce trained models using automated, state-of-the-art algorithms. Built-in evaluation metrics benchmark model performance on test data, while interactive visualizations provide insights into model behavior. Under the hood, the system handles tedious data cleansing, feature engineering, hyperparameter tuning, and model-building steps that otherwise require specialized knowledge. Seamlessly connecting these processes accelerates experiment iteration for users at any skill level. Through a profoundly simplified approach to machine learning, we aim to help visionaries across all domains turn their ideas into practical AI solutions.

The outline of this paper is as follows. Related research is briefed under Section~\ref{related}. The proposed architecture is presented in Section~\ref{approach}. The supporting functions are described in Section~\ref{functions}. Section~\ref{demo} provides a detailed demonstration of two use cases. The conclusion is provided in Section~\ref{conclusion}.

\section{Related Work}\label{related}

Numerous open-source machine learning suites have gained widespread adoption among practitioners and researchers. WEKA (Waikato Environment for Knowledge Analysis)~\cite{frank2005weka} is a widely used open-source machine learning software suite written in Java. It provides a wide range of machine learning algorithms for data mining tasks like data processing, classification, regression, clustering, association rule mining, and visualization. ML algorithms are accessible without having to know anything about programming through a Graphical User Interface. Life science students, however, who are unfamiliar with machine learning algorithms, may find the list of algorithms overwhelming and find it difficult to decide. Our architectural design aims to eliminate the initial apprehension associated with using machine learning applications and to reduce the learning curve by providing a very simple user interface. 

Orange3~\cite{JMLR:demsar13a} is another open-source tool that integrates data visualization, machine learning, and data mining into a user-friendly graphical user interface (GUI). It simplifies exploratory data analysis and interactive visualization with a visual programming approach, making it accessible to users of all skill levels. This feature empowers researchers, data analysts, and educators, especially those with limited programming expertise, to delve into data analysis with more ease than traditional coding requires. Despite its user-friendly design, Orange3 includes advanced functionalities such as SQL integration and a sophisticated system where widgets communicate and pass data through channels, allowing for intricate interactions. However, these advanced features might pose challenges for beginners, particularly in fields like life sciences where individuals may not be familiar with various databases or tools. While Orange3 aims to democratize data analysis, the EndToEndML project takes this a step further by striving to simplify machine learning to its most fundamental components, making it even more accessible for novices without any background in databases or specific analytical tools.

Scikit-learn~\cite{pedregosa2011scikit} leads for its breadth of algorithms and data tools coded in the versatile Python language. It provides a robust set of ML algorithms for supervised and unsupervised problems including classification, regression, clustering, dimensionality reduction, model selection, preprocessing and more. TensorFlow~\cite{tensorflow2015whitepaper} and PyTorch~\cite{NEURIPS20199015} are the other two most popular open-source frameworks for building and deploying deep learning models. TensorFlow provides end-to-end infrastructure from data wrangling to the deployment of deep learning models. For Python-based neural architectures, PyTorch offers strong GPU optimization. They both provide prebuilt libraries for common model architectures and datasets and offer flexibility and control in model architecture design to build complex models. However, expecting beginners to simultaneously learn programming, machine learning theory, and the use of specialized libraries can be a challenging route into this field.

\begin{figure*}[tbp]
\centering
\includegraphics[width=0.98\textwidth]{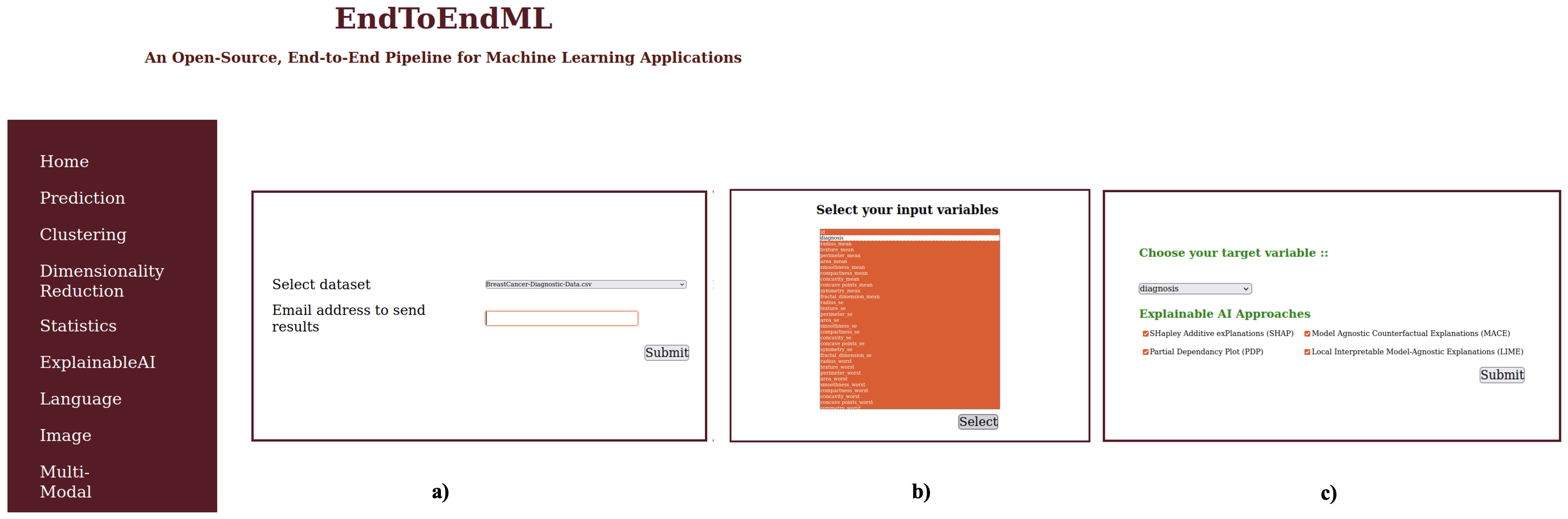}
% This is a *.eps file
\caption{\centering The figure depicts the Graphical User Interface of the EndToEndML tool. It outlines the step-by-step process to select a dataset, choose input features, and specify an output feature for explainable machine learning.}
\label{fig:gui}
\end{figure*}

\section{Architecture} \label{approach}
The EndToEndML pipeline (see fig \ref{fig:1}) aims to create an easy-to-use bioinformatics interface for users, working with large and diverse datasets and with limited programming knowledge. Employing an object-oriented architecture, the graphical frontend exclusively handles intuitive user interaction while encapsulating all computational complexity within the backend. This abstraction focuses interface design on usability rather than functionality. Through this simplified user-facing layer, researchers can configure datasets, parameters and visualize outputs to pilot robust analyses engineered in the backend. Modular software decoupling reinforces flexibility and maintenance of complex data wrangling, machine learning model building, and result generation procedures from user requests.  We designed the system to emphasize an easy user experience rather than technical details under the hood. In this way, all types of users can harness computational power to derive biological insights from their data.  This lets them uncover new discoveries without needing complicated programming.

%The EndToEndML pipeline aims to create a easy-to-use bioinformatics interface for a user, working with large and diverse multiomics datasets and with limited programming knowledge. The approach heavily focuses on the careful designing of the GUI without compromising the robustness of the backend system. The backend and frontend software architecture pattern developed, is based on Object Oriented Programming (OOP) model where the front-end is responsible for the two-way interaction with the users while the back-end handles data pre-processing, model building, training and generating results.

\subsection{Frontend}
Since the focus of EndToEndML was its user-friendliness, we followed a minimalistic approach for the frontend. It follows a streamlined sequence of interfaces from the data upload window all the way through the results window, with options for customization graphically. The final design of the GUI is shown in Figure \ref{fig:gui}. The EndToEndML interface guides users through an intuitive workflow to extract insights from their data. First, users select their desired analysis type such as prediction, clustering, or visualization based on data modalities and use cases (Step a). The system provides an option to select the input data in the case of tabular datasets, or a folder in the case of language, vision, or multi-modal verifications. Recognizing computational runtimes, email notifications allow users to review completed results asynchronously. Minimal configuration is required for initial runs. Users need only specify target variables for prediction and/or input variables for supervised/unsupervised learning from a basic parameters list (Steps b and c). Default preprocessing and validation options apply automated best practices for streamlined analysis. Advanced configuration exposes tuning and algorithm selection. The preprocessing selection could be left untouched by beginners, since our system will choose all the appropriate processing algorithms for the selected dataset automatically. In the following steps, machine learning algorithms will be run sequentially, with the results emailed back to the user.

\subsection{Backend}
\begin{figure}[htbp]
\centering
\includegraphics[width=8 cm]{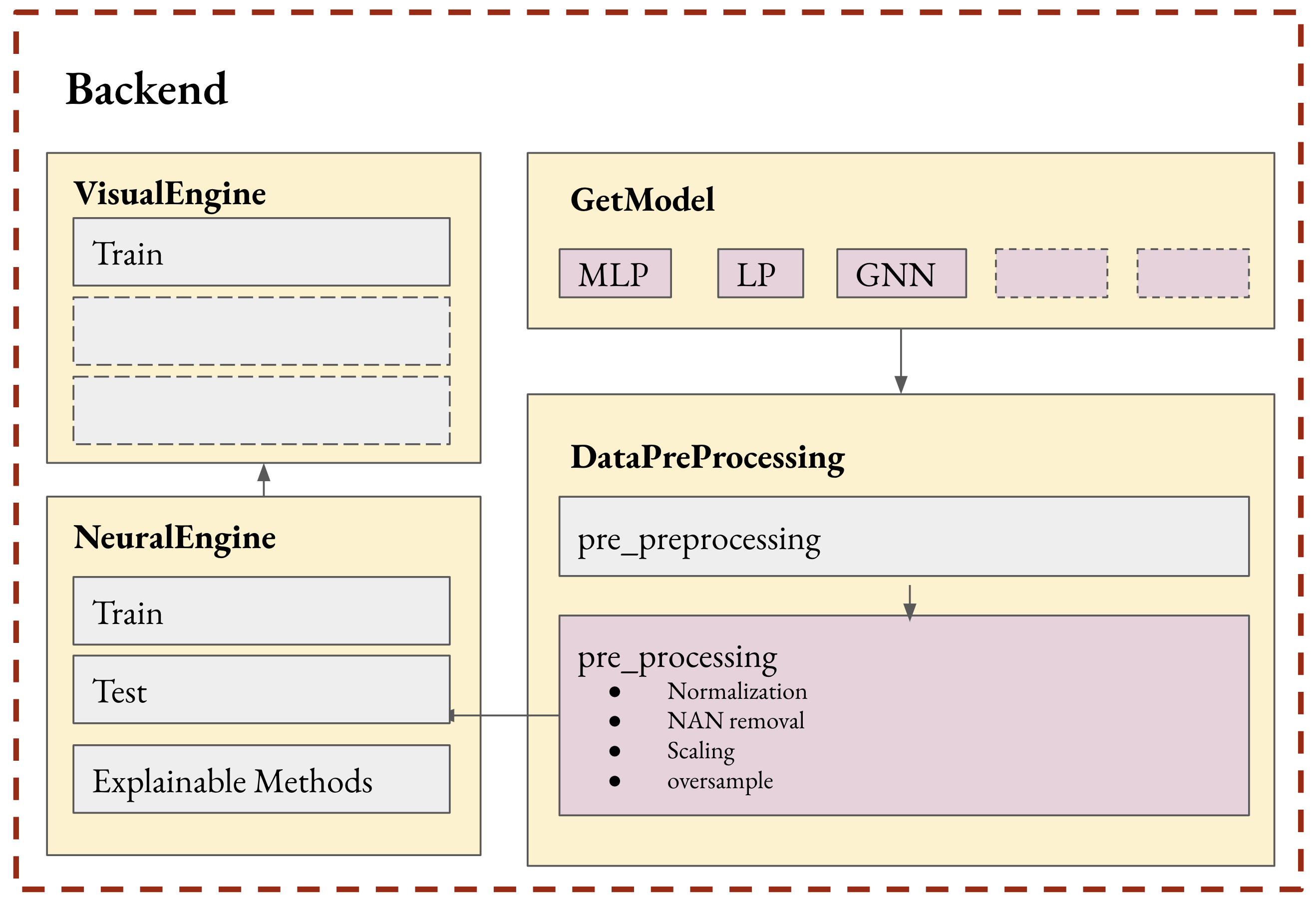}% This is a *.eps file
\caption{\centering Backend Architecture}
\label{fig:2}
\end{figure}

One of the major challenges when developing the backend system for EndToEndML was its complexity. As a single machine learning model can be challenging in itself, while each dataset has its own characteristics, running multiple machine learning and deep learning models on a single dataset added a great deal of complexity. With the use of Object-Oriented Programming (OOP), this problem was mitigated to some extent. During the development stage, a custom error handler object was created and was constantly updated to handle the unique errors that were encountered. An overview of the backend system is shown in figure \ref{fig:2}. The backend system is supported with 4 object classes: DataHandler, ModelEngine, NeuralEngine, and VisualEngine. Each object class is discussed in the following section (see algorithm \ref{alg:backend} for an overall process flow).

\SetKwComment{Comment}{/*}{*/}

\begin{algorithm}
\DontPrintSemicolon
\caption{General backend structure}\label{alg:backend}
\KwData{$Tabular$, $Image$, $Language$, or $Visio-Lingual\ data$}
\KwResult{$Logs$, $graphical\ outputs$}\;

$data \gets Data\ Handler.DataPrePreProcessing(Raw\ Data)$
\Comment*[r]{\textcolor{blue}{Reading from file, removing NaN}}\;

$processed\_data \gets Data\ Handler.DataPreProcessing(data)$
\Comment*[r]{\textcolor{blue}{Scaling, sampling}}\;

$algorithm \gets ModelEngine.GetModel()$
\Comment*[r]{\textcolor{blue}{prediction, clustering}}\;

$model \gets NeuralEngine()$
\Comment*[r]{\textcolor{blue}{training, evaluation}}\;

$VisualEngine()$
\Comment*[r]{\textcolor{blue}{logging, graphs}}\;

\end{algorithm}

\subsubsection{Data Handler} 

DataHandler reads a variety of file formats while transforming heterogeneous data into unified representations suitable for machine learning. As part of the data handler, we have included two data conditioning blocks: DataPrePreProcessing and DataPreProcessing. With DataPrePreProcessing, tabular features of categorical, continuous, and binary types are parsed, recognized and then standardized into homogeneous arrays of numerical values, while DataPreProcessing includes methods for normalizing, standardizing, and oversampling the results. Integrated transformation capabilities allow normalization, rescaling, imputation, and sampling augmentation to rectify distributions statistically and minimize distortions algorithmically. 

\paragraph{Pre-Preprocessing}
The data pre-preprocessing is an interesting as well as the most critical problem in the project. Since the program handles data of diverse formats through a single channel, the pre-preprocessing step is important for standardizing them into a robustly designed data structure (see algorithm \ref{datapreprocess}). An example of this step in the microbiome data context would be to unify the data, which is split in multiple sheets, into a single pandas DataFrame variable. The module also addresses potential typos and irrelevant lines in data which are typically present in a software API-generated file.

\begin{algorithm}[h]

\SetKwFunction{FRead}{ReadData}
\SetKwFunction{Ftarget}{FunctionAnalysis}

\SetKwFunction{Finputs}{ProcessX}

\SetKwFunction{Foutputs}{ProcessY}

\SetKwFunction{Ftransform}{TransformX}

\SetKwFunction{Fjb}{StatMethods}

% Write Function with word ``Def''

  \SetKwProg{Fn}{Function}{:}{\KwRet}
  \Fn{\FRead}{
        Read from CSV, Image,..
       
  }

\SetKwProg{Fn}{Function}{:}{\KwRet}
  \Fn{\Ftarget}{
    Determine target function;
       
  }

\SetKwProg{Fn}{Function}{:}{\KwRet}
  \Fn{\Finputs}{
    Extract Inputs;
       
  }

  \SetKwProg{Fn}{Function}{:}{\KwRet}
  \Fn{\Foutputs}{
    Extract Targets;       
  }

  \SetKwProg{Fn}{Function}{:}{\KwRet}
  \Fn{\Ftransform}{
 Transform Data;    
  }

  \SetKwProg{Fn}{Function}{:}{\KwRet}
  \Fn{\Fjb}{
    Statistical Methods;
       
  }
\label{datapreprocess}
\caption{DataPrePreProcessing module}
\end{algorithm}

\paragraph{Pre-processing} During the pre-processing step, as shown in the Algorithm \ref{dataprocess}, appropriate normalization techniques are applied based on the data types in each column; undefined values (NaN) are removed; the data is scaled; and then a random or SMOTE oversampling is performed to account for class imbalances.

\begin{algorithm}[h]

\SetKwFunction{Fscaler}{Scaler}
\SetKwFunction{Fsampling}{OverSampling}

% Write Function with word ``Def''

  \SetKwProg{Fn}{Function}{:}{\KwRet}
  \Fn{\Fscaler}{
        Unit normalization, robust scaling, standard scaling, power transform, quantile transform;
       
  }

\SetKwProg{Fn}{Function}{:}{\KwRet}
  \Fn{\Fsampling}{
    Random, SMOTE oversampling methods;
       
  }

\caption{DataPreProcessing module} \label{dataprocess}
\end{algorithm}

\subsubsection{ModelEngine}
The ModelEngine contains all the machine learning models defined in a library (see Algorithm \ref{model}). A list of models, based on the user's preference, is returned from the getModel function for training. The models are defined based on the dimensions of the input data, and the hyperparameters are set according to preset defaults.

\begin{algorithm}[h]

\SetKwFunction{Fprediction}{Prediction}
\SetKwFunction{Fclustering}{Clustering}
\SetKwFunction{Fdim}{Dimensionality Reduction}
\SetKwFunction{Fexplain}{ExplainableAI}
\SetKwFunction{Fnn}{Neural Network}
% Write Function with word ``Def''

  \SetKwProg{Fn}{Function}{:}{\KwRet}
  \Fn{\Fprediction}{
        Classification and regression algorithms;
       
  }

\SetKwProg{Fn}{Function}{:}{\KwRet}
  \Fn{\Fclustering}{
   Unsupervised clustering algorithms;
       
  }
\SetKwProg{Fn}{Function}{:}{\KwRet}
  \Fn{\Fdim}{
   Popular dimensionality reduction approaches;
       
  }
\SetKwProg{Fn}{Function}{:}{\KwRet}
  \Fn{\Fexplain}{
  ExplainableAI methods;
       
  }

  \SetKwProg{Fn}{Function}{:}{\KwRet}
  \Fn{\Fnn}{
   Multi-model, Neural Network methods;
       
  }
  
\caption{ModelEngine module} \label{model}
\end{algorithm}

\subsubsection{Neural Engine}
This class goes through the standard training steps for a machine learning model. Major features of the program include the Train/Test Split, which randomly divides collected data randomly into a training set and a test set. The next step is to prepare the training data to feed into the selected model, tuning the parameters against the provided examples and enable it to capture the underlying relationships. The evaluation of trained models using unseen test data in terms of appropriate metrics such as accuracy, AUC, and precision. Next, hyperparameter tuning is conducted to optimize evaluation metrics through new training cycles. Finally, the model will be retrained on the entire dataset to achieve the best results. 

\subsubsection{Visual Engine}
The Visual Engine generates all graphical outputs that need to be returned to the user. Additionally, it handles the logging function for the entire architecture. 

\section{Supporting Functions}\label{functions}

Spanning both exploratory and advanced techniques, EndToEndML's initial release features versatile algorithms designed to extract insights from multi-omics datasets. We have already integrated some popular and widely used algorithms; however, some integrations are currently in progress. For traditional modeling, it includes regression strategies such as logistic models, tree-based techniques including random forests and gradient boosting, instance-based algorithms such as k-nearest neighbors, along with unsupervised methods like k-means clustering. For sequencing or temporal data, recurrent and convolutional neural networks come pre-equipped. State-of-the-art transformer architectures are available for language tasks. Through its modular software architecture, newly published or custom methodologies can be smoothly integrated over time. In addition to enabling the immediate performance of a wide range of analyses, this adaptability allows both novice and expert users to evolve alongside the advancements in the field. With an accessible interface that is equipped with both conventional and emerging machine learning techniques, EndToEndML offers multifaceted ML solutions to match the complexity and promise of AI research in the life sciences.

\subsection{Prediction}
Using Python scikit-learn~\cite{pedregosa2011scikit}, our system leverages robust machine learning algorithms to perform predictions. Scikit-learn supplies a versatile set of supervised and unsupervised learning methods for tasks ranging from regression to classification and clustering. It enables efficient data transformations and model training using techniques such as random forests and neural networks, followed by optimized predictions on new data. Our backend pipeline interfaces with scikit-learn estimators to automatically construct ML models custom-fitted to supplied datasets without requiring coding from users. Seamless access to scikit-learn's extensive catalog enables our application to provide accurate forecasts and data-driven insights across industries for front-facing stakeholders. By managing the coding complexities involved in applying ML, our system democratizes these predictive capabilities, making them accessible to users of any technical skill level to meet various business needs.
\paragraph{Linear Regression} Linear regression~\cite{galton1886regression} is a basic and commonly used type of predictive analysis. It is a statistical method to model the relationship between a dependent variable y and one or more independent variables X. This method quantifies the linear correlation between the variables by fitting a best-fit straight line, known as a linear model, to the observed data points.

\paragraph{Logistic Regression} Logistic regression~\cite{cox1958regression} is a classification algorithm used to predict a categorical dependent variable. Logistic regression transforms its output through the logistic sigmoid function, returning probabilities that range between 0 and 1. This function captures the nonlinear relationships present in the data.

\paragraph{Support Vector Machine} SVMs~\cite{cortes1995support} are supervised machine learning models used generally for classification and regression problems. They construct an optimal hyperplane in n-dimensional space that distinctly classifies the data points. This hyperplane is determined based on the support vectors, which are the data points closest to the hyperplane.

\paragraph{k-Nearest Neighbours} The k-Nearest Neighbors (k-NN)~\cite{Mucherino2009} algorithm is a simple, non-parametric supervised learning algorithm that predicts based on the similarity of data points in feature space to training samples. It operates by storing all training samples and classifying new data points based on a similarity measure (e.g., Euclidean distance) to its k nearest neighbors in the training set. A majority vote of the neighbour classes is taken as the prediction.

\paragraph{Naive Bayes} Naive Bayes~\cite{Webb2010} is a simple yet effective supervised classification algorithm that operates under strong feature independence assumptions. Classification is performed by applying Bayes' rule to compute the probabilities of each class given the feature evidence, and then predicting the class with the highest posterior probability.

\paragraph{Random Forest} Random Forest~\cite{ho1995random} is an ensemble supervised learning technique that combines multiple decision tree models to improve prediction accuracy. It works by creating multiple decision trees during training time. Each tree is trained on a random subset of data features. This approach introduces diversity among the trees, enhancing the model's robustness.

\paragraph{Gradient Boosting} Gradient boosting~\cite{friedman2001greedy} produces a predictive model in the form of an ensemble of weak prediction models, typically decision trees. It strategically combines weak models together for improved predictive performance by targeting errors through gradient descent optimization. The algorithm builds the additive model in a forward, stage-wise fashion, and iteratively adds models to provide a more accurate estimate of the response variable. During each iteration, a new weak learner model (decision tree) is trained with the goal of minimizing the loss function (e.g., MSE for regression, log loss for classification) by fitting on the residual errors left over from the aggregation of all prior learner models.

\subsection{Clustering} 

By leveraging the versatile algorithms and metrics available in the scikit-learn Python library~\cite{pedregosa2011scikit}, our framework constructs ML clustering pipelines that facilitate data transformations, model building, and evaluation, all without requiring manual coding.

\paragraph{K-means} K-means clustering is an unsupervised machine learning algorithm that groups an unlabeled dataset into a predefined number of clusters, denoted by k. It works by identifying k centroids, which are points that represent the mean position of all the data points within a cluster. The algorithm repeatedly assigns each data point to the nearest centroid, forms k clusters, and recomputes the centroid of each cluster. The result is the clustering of data into distinct groups based on similarity, achieved without any prior training. 

\paragraph{DBSCAN} Density-Based Spatial Clustering of Applications with Noise~\cite{ester1996density} is a popular density-based clustering algorithm used in unsupervised machine learning. It clusters data points based on local density, rather than on distance from centroids as in k-means. Regions characterized by a high density of points form clusters. DBSCAN is particularly effective for spatial data and can identify clusters of arbitrary shapes.

\paragraph{Agglomerative} Agglomerative clustering~\cite{ZepedaMendoza2013} is a type of hierarchical clustering algorithm that groups data points into a tree of clusters. This algorithm employs a bottom-up approach, starting with each data point in its own cluster. It proceeds by successively merging the most similar, or ``nearest,'' clusters.

\paragraph{Gaussian mixture model} Gaussian Mixture Models (GMMs)~\cite{yang2012robust} are unsupervised clustering algorithms based on multivariate Gaussian distributions. Data points are assumed to be generated from a combination of a finite number of Gaussian distributions with unknown parameters. GMMs offer probabilistic clustering for heterogeneous data by identifying subpopulations within Gaussian distributions.

\subsection{Dimensionality Reduction}

The present design includes Principal Component Analysis (PCA)~\cite{doi:10.1080/14786440109462720} and Kernel PCA~\cite{10.1007/BFb0020217} to reduce dimensionality. In both cases, the scikit-learn API is used as the backend software. PCA is an unsupervised linear dimensionality reduction technique that seeks to project the data along directions of maximum variance called principal components. It works by calculating a covariance matrix of the features and finding its eigenvectors.  Kernel PCA is a nonlinear form of PCA that uses the kernel trick to transform data and uncover nonlinear latent patterns. Data are implicitly mapped into a higher-dimensional space where the patterns become linear. Then, standard linear PCA is performed in the transformed space to reduce dimensions along principal components.

\subsection{Statistics}
This package includes popular statistical methods such as correlation coefficients and distribution verifications. To implement these modules in our program, we use Scipy~\cite{2020SciPy-NMeth}, a Python library that is free, open-source, and designed for scientific and technical computing. 

\paragraph{Correlation Coefficient} 
The correlation coefficient measures the strength and direction of the linear relationship between two continuous variables.  The Pearson correlation coefficient (r) ~\cite{benesty2009pearson} is a simple and straightforward to calculate and interpret degree of linear association. It utilizes standardized data values to identify correlation ranging from -1 to 1. When the value is 0, there is no relationship. Statistical correlations are expressed as 1 for complete positive correlations and -1 for complete negative correlations. It assesses the linear dependence between samples of two variables, X and Y, by quantifying tightness of data scatter around the regression line. It is calculated by finding the covariance between standardized versions of X and Y divided by their standard deviations. The Spearman's rank correlation coefficient~\cite{zar2005spearman} is a nonparametric measure quantifying the monotonic relationship between two variables. It assesses monotonic relationships, where variables change together but not necessarily in a linear fashion, by converting values to ranks and then calculating the Pearson coefficient on these ranks. Kendall's tau~\cite{kendall1945treatment} is a non-parametric statistical metric measuring the ordinal association between two variables. Rather than using ranks, it measures correlation by observing concordance and discordance of pairs of observations. 
\subsection{Explainable AI}

Explainable AI (XAI) refers to AI systems capable of explaining their reasoning, rationale, and decision-making processes to human users in an understandable manner. XAI is becoming increasingly important as AI systems are deployed in sensitive domains such as healthcare, finance, and criminal justice, where transparency in how they arrive at conclusions is critical. Utilizing the OmniXAI~\cite{Yang2022OmniXAIAL} libraries, we have incorporated several of the most popular explainable AI architectures.

\paragraph{Local Interpretable Model-agnostic Explanations} LIME~\cite{ribeiro2016should} is a popular XAI technique that explains the predictions of any machine learning model by approximating it locally with an interpretable model. It aims to explain individual predictions of complex black box ML models, such as deep neural networks, by fitting simple linear models that approximate the model behavior locally around the prediction. It generates synthetic data samples surrounding the data point to be explained, obtains predictions for these samples, and fits a simple linear model, such as linear regression or a decision tree, to this dataset.

\paragraph{Model Agnostic Counterfactual Explanations} MACE~\cite{Yang2022MACEAE} is another technique for providing explanations of machine learning model predictions. MACE generates counterfactual explanations to explain model predictions. Counterfactuals are examples generated by introducing minimal changes to the input features, thereby altering the model’s prediction from the original output. It aims to identify causal relationships between input variables and model outputs within a localized region to offer intuitive and actionable explanations.

\paragraph{SHapley Additive exPlanations} SHAP~\cite{Lundberg2017AUA} explains the prediction of machine learning models by computing Shapley values from game theory to determine feature contributions. It assigns to each feature a value that represents its impact on the prediction via an explainable additive feature attribution method. Features with high absolute SHAP values contribute predominantly to differences between predictions. The sign of a SHAP value indicates whether the feature increased or decreased the prediction.

\paragraph{Partial Dependence Plot} PDP~\cite{pdp} is a visualization technique used in XAI to understand how machine learning models respond to changes in input features. PDPs illustrate the marginal effect that one or two features exert on the predicted outcome of a machine learning model by averaging out the influences of all other features.

\subsection{Language}
 Bidirectional Encoder Representations from Transformers (BERT)~\cite{kenton2019bert} have been incorporated into our architecture to facilitate language processing functions with the support of the PyTorch deep learning library~\cite{paszke2019pytorch}. BERT utilizes transformers, reading text both sequentially and in reverse order, to learn contextual relationships in both directions.
 
\subsection{Image} The Vision Transformer (ViT)~\cite{dosovitskiy2020image} is incorporated using the Keras neural network library~\cite{chollet2015keras} for performing image recognition tasks. ViT employs a pure transformer architecture, effectively modeling global contexts and dependencies among image regions.

\subsection{Visual question answering (VQA)}
Visual question answering~\cite{antol2015vqa} seeks to mimic the human ability to interpret images and answer natural language questions about visual content. In VQA systems, users pose free-form questions related to the details of a photograph or another form of image presented. Algorithms subsequently analyze pixels and text to automatically generate answers in fluent natural language without human intervention. By integrating computer vision and natural language processing, these systems initially recognize objects, actions, scenes, and attributes within images. Concurrently, natural language understanding modules extract the intent and semantics from the open-ended questions. Through multimodal reasoning, linguistic and visual comprehensions are unified to ascertain the most appropriate response. Our current visual question answering (VQA) framework and models employ PyTorch deep learning library~\cite{paszke2019pytorch} as the underlying implementation framework.

\section{Practical Usecase Demonstration} \label{demo}
\begin{figure}[htbp]
\centering
\includegraphics[width=8.75cm]{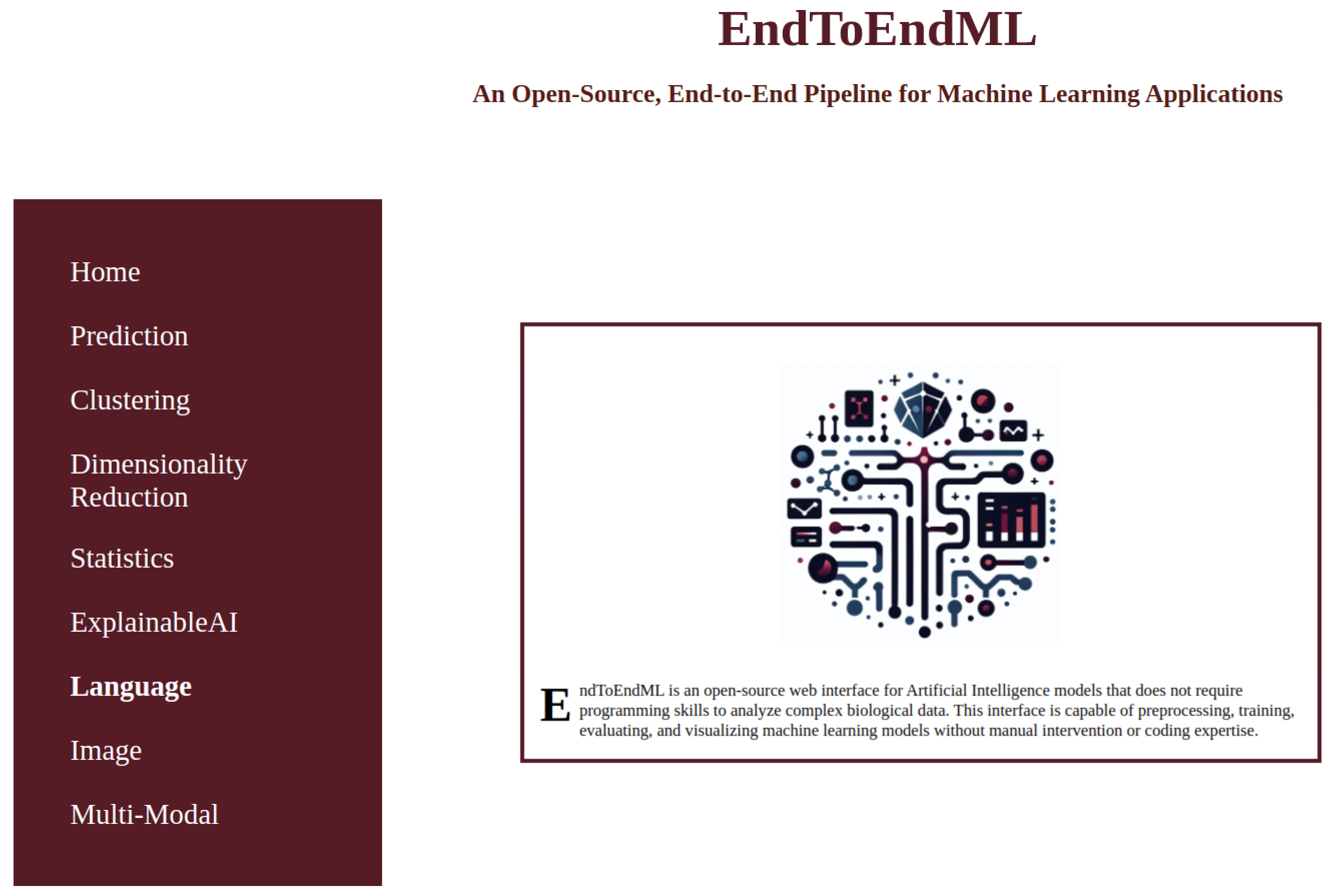}% This is a *.eps file
\caption{\centering Frontend graphical user interface}
\label{frontend}
\end{figure}

To demonstrate the functionality and simplicity of our approach, we present two examples using datasets related to biological sciences.  Figure~\ref{frontend} illustrates the frontend of our graphical user interface. The user is required to select the appropriate option: 'Language' to generate a language model based on their dataset, or 'Multi-Modal' for multi-sensor applications such as Visual Question Answering (VQA).

\subsection{Language}
For this demonstration, we selected a Kaggle dataset~\cite{kaggle_speech} entitled ``Medical Speech, Transcription, and Intent,'' comprising English language transcriptions and common medical symptoms. Our objective is to generate a language model using this medical data and provide a trained model that users can later utilize to develop a medical chatbot. Through this process, we introduce users to the language-based neural network facets of machine learning.

\begin{figure}[htbp]
\centering
\includegraphics[width=0.45\textwidth]{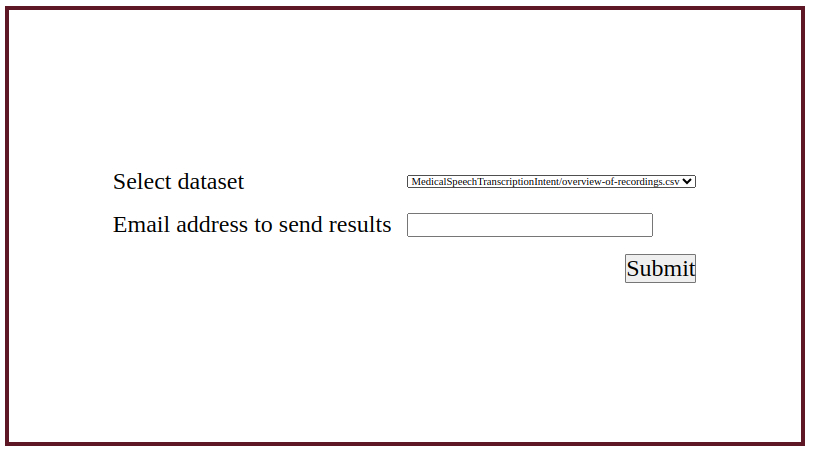}

\caption{\centering Dataset selection}
\label{dataset}
\end{figure}

Upon selecting the ``Language'' tab from the EndToEndML frontend, our tool directs users to a simple webpage where they can choose their dataset (Figure ~\ref{dataset}), which should be downloaded and placed in a pre-selected folder, from a dropdown menu. Additionally, we offer users the option to enter their email address, enabling them to receive a notification that includes the results and the location of the saved result folder for future reference.

\begin{figure}[htbp]
\centering
\includegraphics[width=0.45\textwidth]{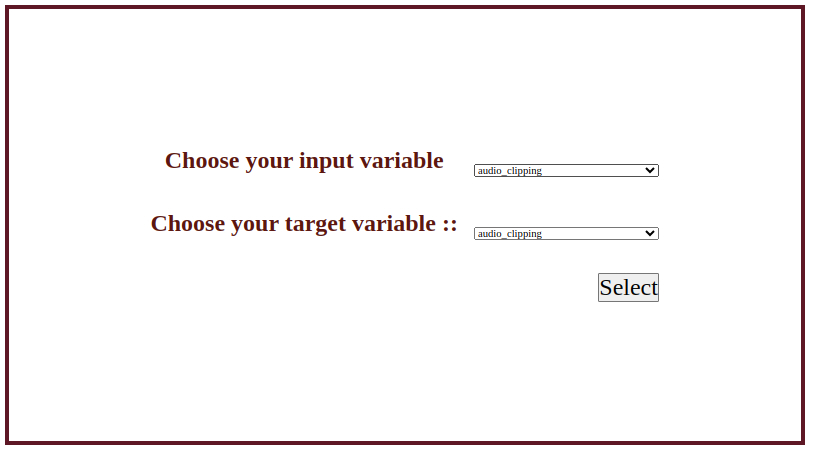}

\caption{\centering Feature selection}
\label{lang_variables}
\end{figure}

We selected a CSV file containing multiple medical variables as features. The subsequent step (Figure~\ref{lang_variables}) involves selecting the input variable, which consists of a sequence of text data that can be used to build a language model and generate a chatbot. Furthermore, it is necessary to select the target feature from the CSV file, which stores medical symptom information corresponding to the medical transcription (input feature).

\begin{figure}[htbp]
\centering
\includegraphics[width=0.45\textwidth]{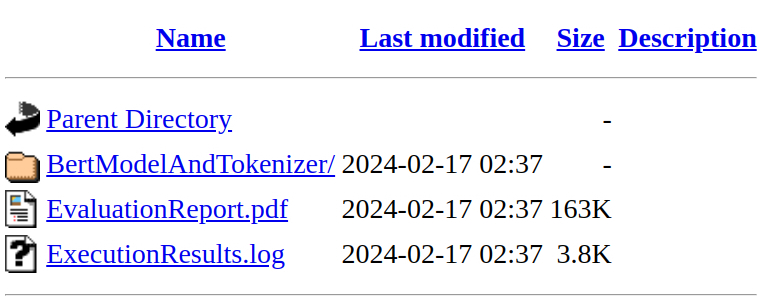}
\caption{\centering Results}
\label{lang_result}
\end{figure}

Upon the selection of relevant features, our backend Application Programming Interface (API), constructed with the Python programming language~\cite{van1995python}, initiates the process of constructing a language model. This model is generated by leveraging the sequence of medical transcriptions and their corresponding medical symptoms present in the dataset. The learning phase may extend over several hours, depending on the server employed and the size of the dataset. Upon the conclusion of the learning process, our Graphical User Interface (GUI) displays an evaluation report in PDF format (Figure~\ref{lang_result}). Additionally, it stores the trained language model and other essential files required for the future development of a chatbot application.

\begin{figure}[htbp]
\centering
\includegraphics[width=0.4\textwidth, angle=0.4]{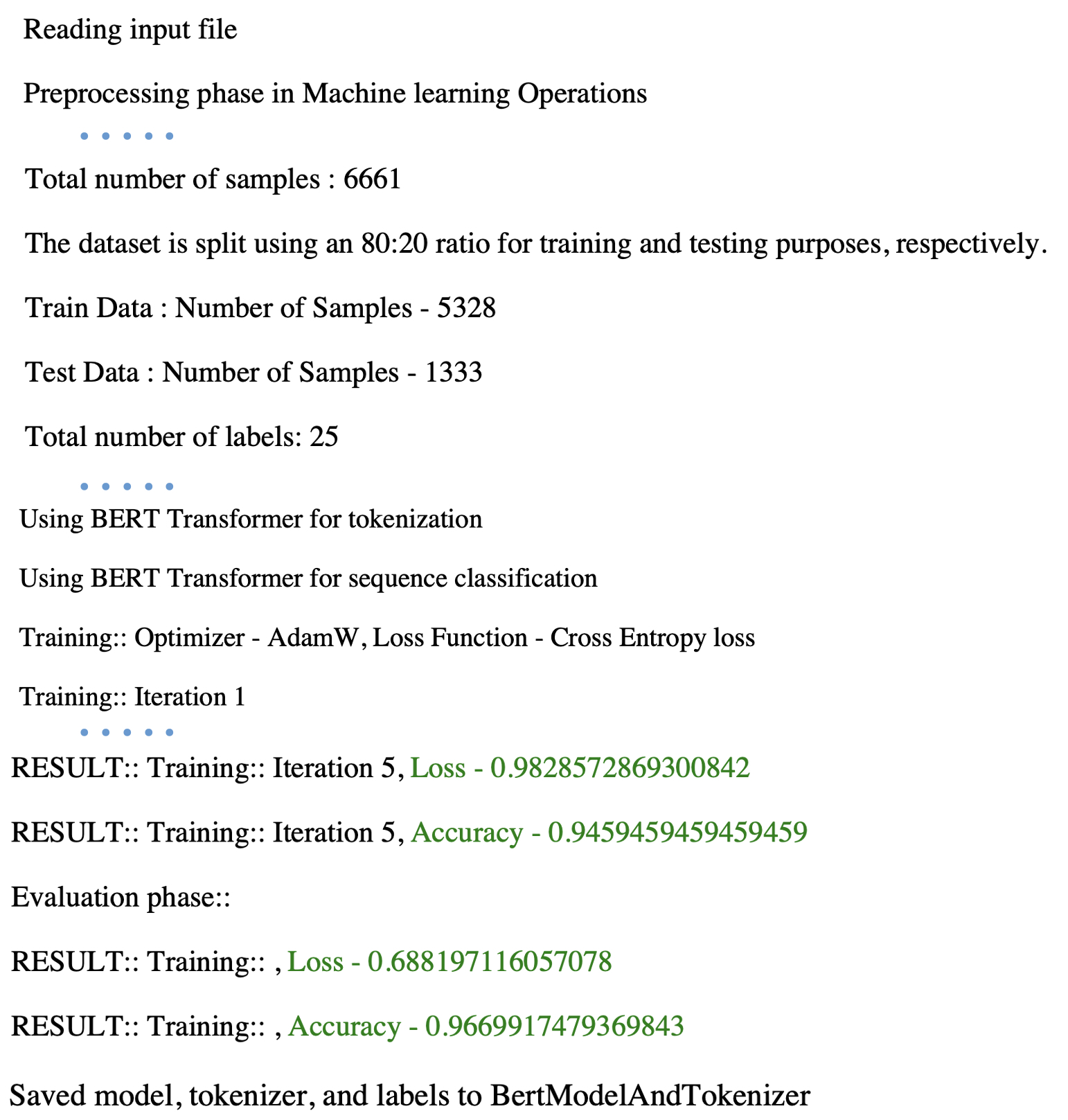}

\caption{\centering Language model evaluation Report}
\label{lang_report}
\end{figure}

Figure~\ref{lang_report} illustrates a sample evaluation report. Initially, it details the dataset, including the number of samples and the language sequences corresponding to each medical symptom. Subsequently, it provides information about the training phase, detailing the training parameters employed in the learning process. Currently, our framework employs the Bidirectional Encoder Representations from Transformers (BERT)~\cite{kenton2019bert} tokenizer and the BERT language model for the learning process. It then displays the batch-wise training loss and accuracy. Upon completion of the learning phase, the report presents the total training loss and accuracy.

Subsequently, the process transitions to the evaluation phase, during which the test data are assessed using the learned model, and both the classification loss and accuracy are reported. Furthermore, this phase includes a classification report generated with the scikit-learn library. Finally, our software saves all learned model artifacts, including the tokenizer and label details, facilitating their subsequent utilization.

\subsection{Visual Question Answering}

For the multi-modal learning approach, we utilized a medical visual question answering (VQA) dataset~\cite{he2020pathvqa}. In this setup, an AI system is presented with a pathology image alongside a question and is tasked with generating the appropriate answer. This dataset comprises a range of open-ended questions along with their corresponding answers, all of which are based on pathology images.
\begin{figure}[htbp]
\centering
\includegraphics[width=0.4\textwidth,angle=0.8]{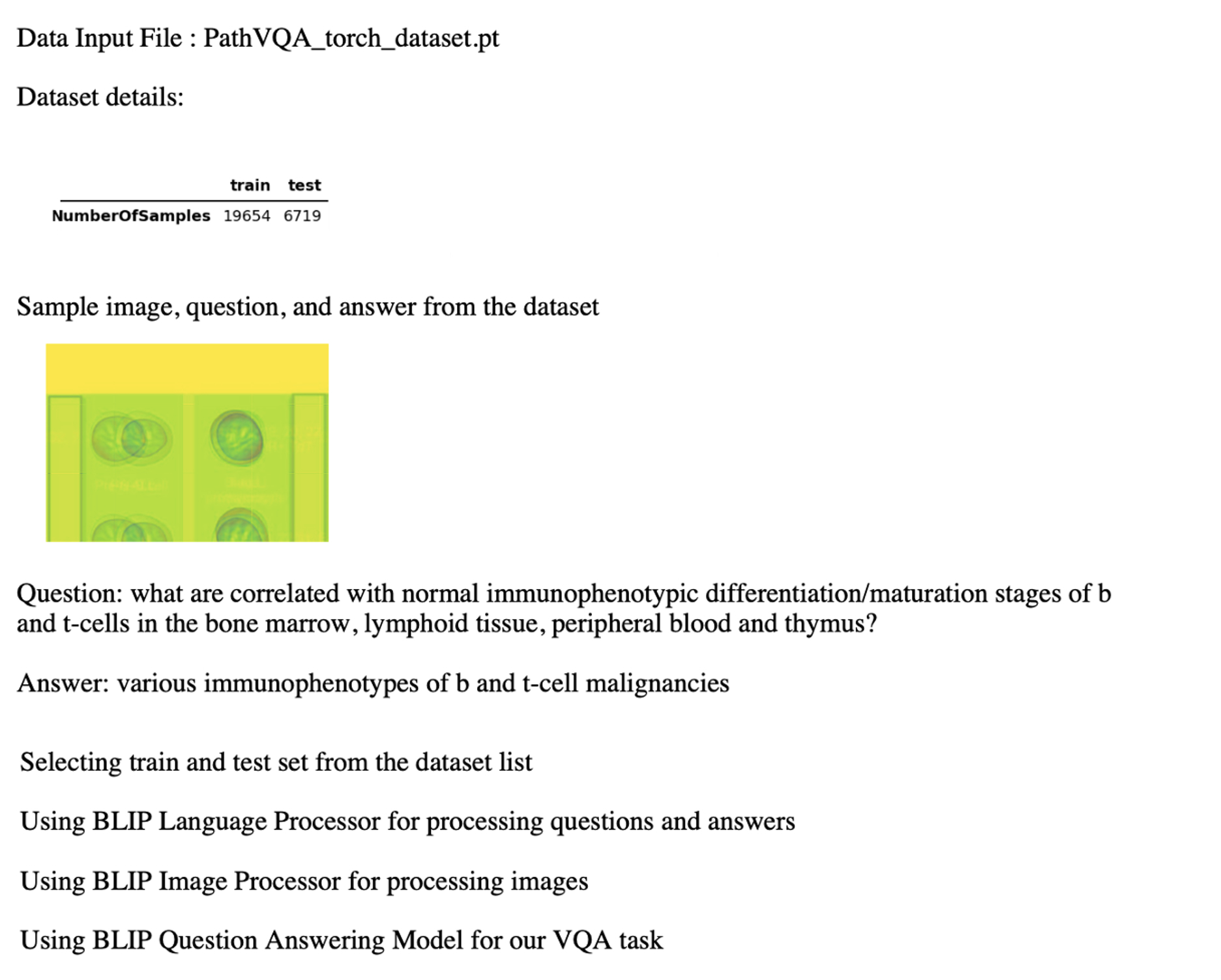}
\caption{\centering Visual Question Answering (VQA) training parameters}
\label{vqa_initial}
\end{figure}
In this scenario, users initiate the process by selecting the ``Multi-Modal'' option within the application's interface (Figure~\ref{frontend}), subsequently guiding them to the dataset selection stage (Figure~\ref{dataset}). At this point, users can select a specific dataset folder for their project. The dataset in question is characterized by three elements: an image, a question related to that image, and the answer to the question. Our goal is to develop a multi-modal visio-language model that is capable of accurately answering questions presented alongside an image during the evaluation phase.

\begin{figure}[htbp]
\centering
\includegraphics[width=0.4\textwidth,angle=0.4]{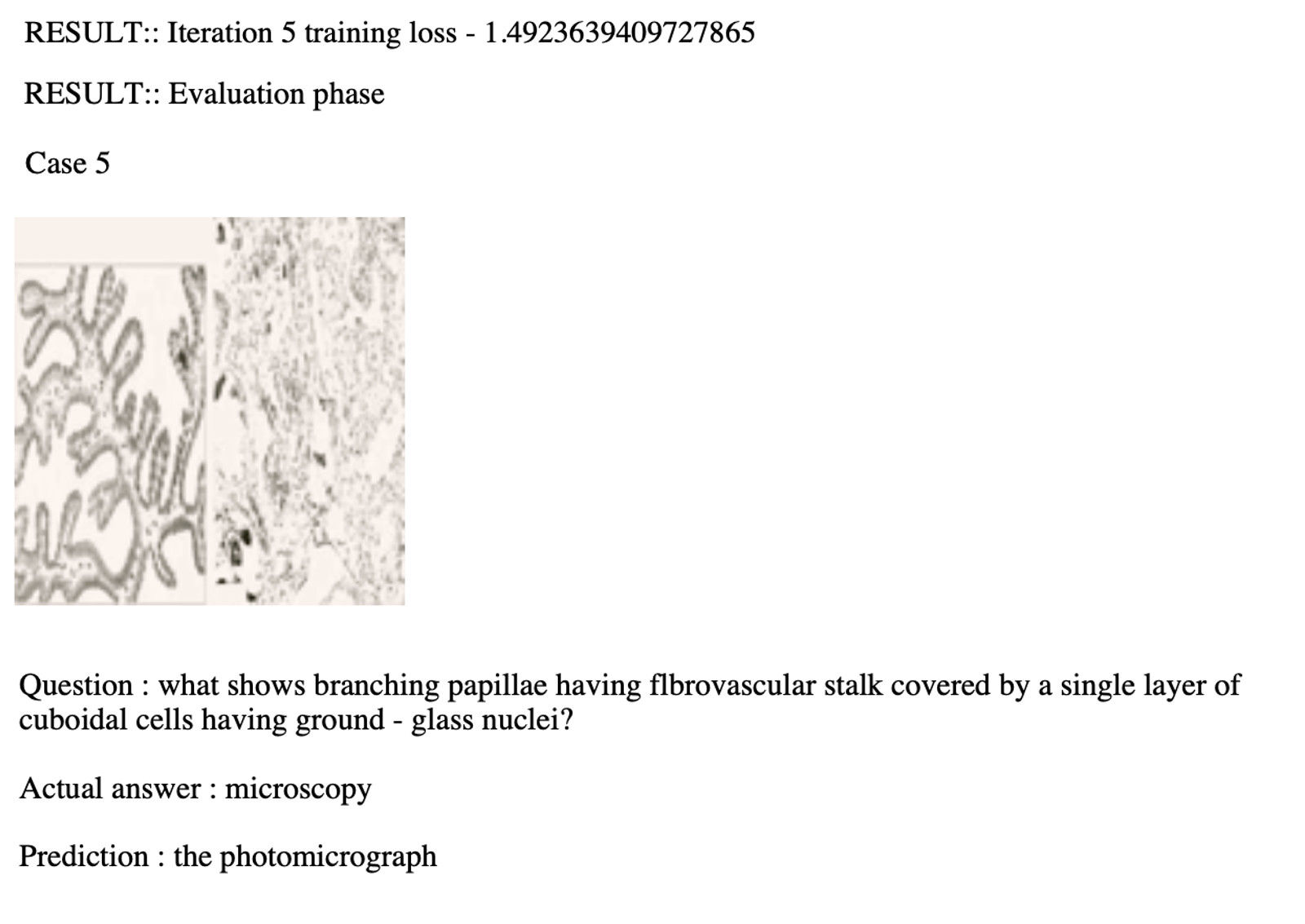}
\caption{\centering VQA model training loss and evaluation performance. }
\label{vqa_results}
\end{figure}

Upon activation of our VQA (Visual Question Answering) model, the system begins to analyze dataset features, alongside training and testing parameters. We integrate the Bootstrapping Language-Image Pre-training (BLIP)~\cite{https://doi.org/10.48550/arxiv.2201.12086} image processor and the BLIP language processor to manage our image and language data, respectively. Additionally, we employ the BLIP VQA model, utilizing Hugging Face transformers~\cite{wolf2020transformers} to develop our VQA framework. Comprehensive details of this process (Figure~\ref{vqa_initial}), including outcomes and evaluations, are documented in a report provided to the user upon completion of the execution. Subsequent to the learning process, we report (Figure~\ref{vqa_results}) the training and evaluation loss as key evaluation metrics. Following this, we evaluate the performance of the trained Visual Question Answering (VQA) model using our test dataset, which comprises images and their corresponding questions.

\section{Conclusion} \label{conclusion}

We present a user-friendly and intuitive Graphical User Interface (GUI) accessible to beginners in the life sciences domain, enabling them to adopt machine learning techniques and obtain a comprehensive evaluation report. This report includes detailed information on the dataset, training parameters, evaluation metrics, results, and performance evaluations. Currently, our EndToEndML tool integrates traditional machine learning methods, renowned explainable AI techniques, and learning architectures for language, visual, and multimodal data. In the future, we plan to incorporate advanced language and image models into our training methodologies. Furthermore, we aim to integrate bioinformatics-related tools into this framework, thereby establishing a unified bio-neural network architecture for learners.

\bibliography{references}
\bibliographystyle{IEEEtran}

\end{document}